\documentclass{article}
\usepackage{spconf,amsmath,graphicx}

\usepackage{cite}
\usepackage{amsmath,amssymb,amsfonts}
\usepackage{algorithmicx}
\usepackage{algorithm}
\usepackage{graphicx}
\usepackage{textcomp}
\usepackage{xcolor}
\usepackage{hyperref}
\usepackage{algpseudocode}
\algtext*{EndIf}
\algtext*{EndFor}
\usepackage{multirow}
\usepackage{multicol}
\usepackage{booktabs}
\usepackage{color}
\usepackage{caption}
\usepackage{subcaption}
\usepackage[inline]{enumitem}

\setlist{nosep, leftmargin=14pt}

\usepackage{mwe} 

\title{Adapting Foundation Models for Few-Shot Medical Image Segmentation: Actively and Sequentially}
\name{Jingyun Yang, Guoqing Zhang, Jingge Wang, Yang Li\thanks{Correspondence: yangjy20@mails.tsinghua.edu.cn}
\thanks{All authors are from the Shenzhen Key Laboratory of Ubiquitous Data Enabling, SIGS, Tsinghua. This work is supported in part by the Natural Science Foundation of China (Grant 62371270)  and  Shenzhen Key Laboratory of Ubiquitous Data Enabling (No.ZDSYS20220527171406015)}}

\address{Tsinghua Shenzhen International Graduate School, Tsinghua University}

\begin{document}
%
\maketitle
\begin{abstract}
Recent advances in foundation models have brought promising results in computer vision, including medical image segmentation.
Fine-tuning foundation models on specific low-resource medical tasks has become a standard practice.
However, ensuring reliable and robust model adaptation when the target task has a large domain gap and few annotated samples remains a challenge. 
Previous few-shot domain adaptation (FSDA) methods seek to bridge the distribution gap between source and target domains by utilizing auxiliary data.
The selection and scheduling of auxiliaries are often based on heuristics, which can easily cause negative transfer.  
In this work, we propose an Active and Sequential domain AdaPtation (ASAP) framework for dynamic auxiliary dataset selection in FSDA.
We formulate FSDA as a multi-armed bandit problem and derive an efficient reward function to prioritize training on auxiliary datasets that align closely with the target task, through a single-round fine-tuning.
Empirical validation on diverse medical segmentation datasets demonstrates that our method achieves favorable segmentation performance, significantly outperforming the state-of-the-art FSDA methods. 
Code is available at 
\href{https://github.com/techicoco/ASAP}{ASAP}.
\end{abstract}
\vspace{-0.1cm}
\section{Introduction}
\vspace{-0.1cm}
\label{sec:intro}
Recent works like SwinUNet\cite{cao2022swin}, MambaUNet \cite{wang2024mamba} and MONAI \cite{cardoso2022monai} develop medical-tailored foundation models on large-scale medical image datasets. 
Intense interest has emerged in adapting these foundation models for specific medical image analysis tasks. 
However, the generalization capability of foundation models is limited by the large variability in training data, due to complex modalities, intricate anatomical structures, and wide-range object scales in medical images.
Therefore, we seek to answer this critical question: how to effectively adapt these foundation models to our desired medical image processing tasks? 

Unlike natural image analysis with large-scale labeled datasets, in medical image analysis, another major challenge is the lack of labeled data, as 
annotating disease-specific medical images is not only time-consuming but also demands 
specialty-oriented skills, leading to the problem of few-shot domain adaptation (FSDA). 
Most solutions to conventional domain adaptation problems either assume access to source data \cite{bermudez2018domain}, which is not always feasible in real-world medical scenarios with various regulatory standards and ethical considerations,
or they require a substantial amount of unlabeled target data to reduce the distribution gap across domains, as seen in unsupervised domain adaptation (UDA) methods  \cite{wu2021unsupervised}.
FSDA, on the other hand, addresses the situation when only a limited number of target examples are available for training, whether labeled or unlabeled.
Previous FSDA methods \cite{gu2019progressive} propose to use intermediate/auxiliary domains to facilitate domain adaptation.
However, this multi-step domain adaptation strategy requires fine-tuning the model twice or more.
In this work, we propose to incorporate auxiliary datasets to solve the FSDA problem in a source-free manner through a single-round fine-tuning.

Training with auxiliary data introduces an inductive bias that helps models capture meaningful representations and reduces the risk of overfitting to spurious correlations \cite{navon2021auxiliary}.
Multi-task learning methods \cite{graham2023one} cannot extend to a large number of tasks because the complexity of the search space will be exponentially explosive \cite{albalak2024improving}.
Other strategies in auxiliary learning and transfer learning hand-pick which auxiliary data to use based on heuristics \cite{yang2021joint} or metrics \cite{yu2020gradient} prior to training, sometimes resulting in sub-optimal outcomes.
Recent dynamic auxiliary learning works \cite{navon2021auxiliary} propose to dynamically combine auxiliary objectives through task or data schedulers, but these methods involve complex and computationally demanding bi-level optimization steps.

To address the above issues, we propose an \textbf{A}ctive and \textbf{S}equential domain \textbf{A}da\textbf{P}tation (ASAP) framework for FSDA.
Using a novel dynamic dataset selection strategy,
the proposed framework prioritizes training on auxiliary datasets with similar solution spaces to the target task in a \textbf{single-round} computational complexity.
Specifically,
we formulate FSDA as a multi-armed bandit problem in active learning \cite{macready1998bandit} and relate the set of auxiliary datasets to the arms.
We introduce the classic trace upper confidence bound algorithm \cite{auer2002finite} to solve the multi-armed bandit problem.
By balancing the trade-off between the exploration of unobserved arms and the exploitation of high-reward arms, we actively and sequentially select the auxiliary dataset at each turn, maximizing their benefits.
The reward functions we design add minimal memory and computational overhead.

Extensive experiments on three public medical datasets validate the effectiveness of our proposed ASAP framework.
We efficiently adapt pre-trained 
UNet \cite{ronneberger2015u}, SwinUNet \cite{cao2022swin} and MambaUNet \cite{wang2024mamba} from Flemme
\cite{zhang2024flemme}, 
a flexible and modular learning medical platform, for various target medical image segmentation tasks.
Our method outperforms the FSDA auxiliary learning methods with lower computation costs.
Our main contributions are as follows:
\begin{itemize}
    \item \textbf{\textit{An active and sequential domain adaptation framework}}: we propose a novel framework that incorporates auxiliary datasets to effectively adapt foundation models in a single-round fine-tuning for various medical segmentation tasks, optimizing the use of public medical resources.
    \item \textbf{\textit{An exploration-exploitation balanced  FSDA algorithm}}: we design an efficient reward function and successfully apply the multi-armed bandit algorithm to dynamic auxiliary dataset selection through the ASAP framework.
\end{itemize}

\vspace{-0.2cm}
\section{Methodology}
\begin{figure}
    \vspace{-0.1cm}
    \centering
    \includegraphics[width=\linewidth]{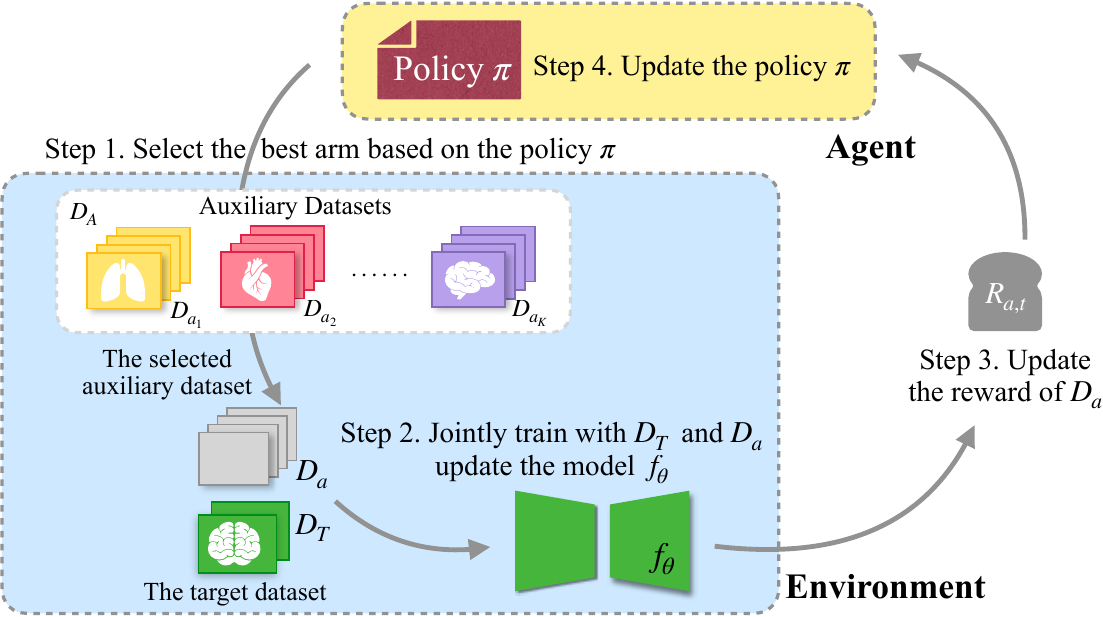}
    \vspace{-0.5cm}
    \caption{Illustration of our active and sequential domain adaptation (ASAP) framework.
    The agent defines the policy $\pi$ that determines which arm to pull. The environment includes the auxiliary data pool $\mathcal{D}_\mathcal{A}$, the target dataset $\mathcal{D}_\mathcal{T}$, and the model $f_\theta$. At each turn $t$, ASAP executes the four shown steps.
    }
    \label{framework}
    \vspace{-0.2cm}
\end{figure}

\vspace{-0.2cm}
In this section, we will elaborate on the proposed active and sequential domain adaptation (ASAP) framework, shown in Fig.~\ref{framework}.
First, we clarify the setting of few-shot domain adaptation with auxiliary datasets.
Then we formulate it as a multi-armed bandit (MAB) problem and describe how we solve it.

\vspace{-0.2cm}
\subsection{Problem definition}
\vspace{-0.1cm}
For domain adaptation problems, the network is usually first trained on an adequate source domain dataset $\mathcal{D}_\mathcal{S}$. 
We denote the pre-trained source model as $\Theta_s$.
Given a small quantity of data belonging to a target domain dataset $\mathcal{D}_\mathcal{T}=\{(x_i^t,y_i^t)\}_{i=1}^{m}$, the goal is to adapt $\Theta_s$ to achieve high performance on $\mathcal{D}_\mathcal{T}$ with access to a set of available auxiliary datasets $\mathcal{D}_\mathcal{A} = \{\mathcal{D}_{a_1},\mathcal{D}_{a_2},...,\mathcal{D}_{a_K}\}$.
For all $a \in A$, $\mathcal{D}_a$ is an individual auxiliary dataset.

In this work, we formulate the auxiliary data selection problem in FSDA as a Markov decision process by adopting the multi-armed bandit (MAB) setting \cite{macready1998bandit}.
MAB has been used in sequential experiment design in active learning,
where the goal is to sequentially choose experiments to perform with the aim of maximizing some outcomes.
The MAB learning paradigm involves 
an agent interacts with an environment over $N$ turns by following a policy $\pi$.
In our work, the environment consists of the target dataset $\mathcal{D}_\mathcal{T}$, the set of auxiliary datasets $\mathcal{D}_\mathcal{A}$, and the model $f_\theta$.
The agent learns a policy $\pi$ that defines the selection strategy over all $\mathcal{D}_a\in\mathcal{D}_\mathcal{A}$. 
At each turn $t$, the agent selects one of the environment’s $K$ datasets $\mathcal{D}_a\in\mathcal{D}_\mathcal{A}$
to jointly trained with $\mathcal{D}_\mathcal{T}$.
The environment then updates the model $f_\theta$.
Accordingly, the agent receives a reward $R_{a,t}$ and uses it to update the policy $\pi$.
Rewards for unplayed arms are not observed. 
The goal of the agent is to adopt a policy $\pi$ that selects actions that lead to the largest cumulative reward over $N$ turns, $R=\sum_{t=1}^N R_t$. 
\vspace{-0.45cm}
\subsection{Deriving an efficient reward function}
\vspace{-0.1cm}
To ensure that the decision-making process adds minimal memory and computational overhead,
we derive rewards from the model's intrinsic information and the optimized losses, rather than relying on an external model or metric that requires extra training.
To achieve positive transfer during sequential adaptation, we design the reward with two considerations in mind: positive for model convergence and positive for joint training with the target task.

Formally, at turn $t$ for the auxiliary dataset $\mathcal{D}_a$
the reward of positive for model convergence is defined as:
\vspace{-0.1cm}
\begin{equation}
\label{PM}
    \mathcal{R}^{PM}_{a,t} = -\mathcal{L}_{a,t} = -\mathcal{L}(f_{\theta_t}, \mathcal{D}_a).
\end{equation}
Let $\nabla_{a} = \nabla_\theta \mathcal{L}(f_{\theta_t}, \mathcal{D}_a)$ be the auxiliary dataset gradient and $\nabla_\mathcal{T} =\nabla_\theta\mathcal{L}(f_{\theta_t}, \mathcal{D}_\mathcal{T})$ be the target dataset gradient,
we denote the reward of positive for joint training with $\mathcal{D}_\mathcal{T}$ as:
\vspace{-0.1cm}
\begin{equation}
\vspace{-0.1cm}
\label{PT}
     \mathcal{R}^{PT}_{a,t} = \frac{\nabla_a\cdot\nabla_\mathcal{T}}{||\nabla_a||_2 ||\nabla_\mathcal{T}||_2}.
\end{equation}
Overall, at turn $t$ the reward of the auxiliary dataset $\mathcal{D}_a$ is defined as:
\vspace{-0.3cm}
\begin{equation}
\vspace{-0.1cm}
\label{reward}
    \mathcal{R}_{a,t} = \alpha \mathcal{R}^{PM}_{a,t} + (1 - \alpha) \mathcal{R}^{PT}_{a,t},
\end{equation}
where $\alpha$ is a time-variant weight that decreases with each selection turn.
Considering that $\mathcal{R}_{a,t}$ relies on the loss and gradients, which are intrinsic to the model, $\mathcal{R}_{a,t}$ is naturally training-efficient.
\vspace{-0.2cm}
\subsection{The decision-making policy}
\vspace{-0.2cm}
To optimize a decision-making policy, we propose to adopt a trace upper confidence bound (UCB) algorithm \cite{auer2002finite} 
where the agent greedily selects arms according to their upper confidence bound. 
Formally, after pulling the arm $a$ at turn $t$, the agent receives a observed reward $R_{a,t}$ and then calculate the estimated mean reward as:
\vspace{-0.1cm}
\begin{equation}
\vspace{-0.1cm}
    \hat{R}_{a} = (1-\beta) \hat{R}_{a} + \beta R_{a,t},
\vspace{-0.1cm}
\end{equation}
where $\beta$ is the smoothing factor \cite{wei2021nonstationary}.
Accordingly, we define the upper confidence bound based on Hoeffding’s inequality \cite{auer2002finite} for arm $a$ at turn $t$ being played $n_a$ times:
\vspace{-0.2cm}
 \begin{equation}
 \vspace{-0.2cm}
    UCB_{a,t} =
    \begin{cases}
        \infty,&\text{if } n_a = 0 \\
        \hat{R}_{a} + \sqrt{\frac{2\ln t}{n_a}},&\text{otherwise}.
    \end{cases}
\end{equation}
This allows us to balance the exploitation of arms with a high predicted reward and the exploration of areas with high uncertainty.
The proposed algorithm is shown in Algorithm.~\ref{ucb}.
\vspace{-0.2cm}
\begin{algorithm}
  \caption{The MAB decision-making policy}
  \label{ucb}
  \begin{algorithmic}[1]
    \Require $\mathcal{D}_\mathcal{A}, \mathcal{D}_\mathcal{T}$: Auxiliary and target datasets
    \Require $f_\theta$: Parameterized model
    \Require $\alpha,\beta$: Decaying  and smoothing factors
    \State \textbf{Initialize} $f_{\theta_0}=\Theta_s$
    \State \textbf{Initialize} the information of each arm $a \in A$\\
    $\begin{array}{ll}
        \forall a \in A: & n_a = 1, \\
        &\nabla_{a} = \nabla_\theta \mathcal{L}(f_{\theta_0}, \mathcal{D}_a),
        \nabla_\mathcal{T} =\nabla_\theta\mathcal{L}(f_{\theta_0}, \mathcal{D}_\mathcal{T}),\\
        & \mathcal{R}^{PM}_{a,0} = -\mathcal{L}(f_{\theta_0},\mathcal{D}_a),
        \mathcal{R}^{PT}_{a,0} = 
        \cos(\nabla_{a}, \nabla_\mathcal{T}),\\
        & \hat{\mathcal{R}}_{a} = 0.5 \mathcal{R}^{PM}_{a,0} + 0.5 \mathcal{R}^{PT}_{a,0}
        \end{array} $
    \For{$t = 1, 2, \dots, N$}
        \State Calculate the upper confidence bound for each arm
        \State $a^* = \arg\max_{a \in \mathcal{A}} \left(\hat{\mathcal{R}}_{a} + \sqrt{\frac{2 \ln t}{n_a}}\right)$
        \State Select the auxiliary dataset $\mathcal{D}_{a^*}$
        \State $n_{a^*} \leftarrow n_{a^*} + 1$

        \State $\nabla_\mathcal{T} \leftarrow\nabla_\theta\mathcal{L}(f_{\theta_{t-1}}, \mathcal{D}_\mathcal{T})$ 
        \State $\nabla_{a^*} \leftarrow \nabla_\theta \mathcal{L}(f_{\theta_{t-1}}, \mathcal{D}_{a^*})$
        \State Update model parameters w.r.t. $\nabla_\mathcal{T} + \nabla_{a^*}$
        \State Update the reward of the pulled arm
        \State $R_{a,t} = \alpha R^{PM}_{a,t} + (1-\alpha) R^{PT}_{a,t}$
        \State $\hat{R}_{a} \leftarrow (1 - \beta)\hat{R}_{a} + \beta R_{a,t}$
        \State Release memory $\nabla_{a^*} \leftarrow 0$
    \EndFor
    \State \textbf{end for}  \\
    \Return $f_\theta$
  \end{algorithmic}
  \vspace{-0.1cm}
\end{algorithm}
\vspace{-0.3cm}

\vspace{-0.5cm}
\section{Experiments and Results}
\vspace{-0.2cm}
To showcase the flexibility of our ASAP framework, we conduct extensive experiments on MRI and CT datasets covering various modalities and anatomical regions.
\vspace{-0.2cm}
\subsection{Datasets and Implementation Details}
\vspace{-0.1cm}
For MRI experiments, 
we construct the auxiliary datasets pool based on FeTS 2022 \cite{pati2021federated} 
(brain tumor segmentation) and iSeg2019 \cite{sun2021multi} (brain tissue segmentation). The auxiliary task pool consists of 30 datasets, each with a sample size exceeding 30.
For the target datasets, we use two brain 3D MRI segmentation datasets:
the periventricular leukomalacia (PVL) dataset \cite{yang2021joint}, characterized by tissue reduction in periventricular and manually delineated on each slice of the patient’s T2 MRI images, and the White Matter Hyperintensity (WMH) dataset \cite{kuijf2019standardized}
which segments white matter hyperintensities on FLAIR MRI images.
For CT experiments, we construct 30 datasets from TotalSegmentator (TOS) \cite{wasserthal2023totalsegmentator} as the auxiliary datasets, based on label diversity and density.
TOS is a whole-body-segmented 3D CT dataset that contains 117 main default tasks.
For the target datasets, we experiment with vessel and liver segmentation tasks from MSD, a benchmark 3D CT dataset \cite{antonelli2022medical}.
For each auxiliary dataset, we use at most 30 training examples.
For each target task, all the experiments are conducted under the 1-way 3-shot scenario using 5-fold cross-validation.
We experimented with 5, 3, and 2 target samples, and found that using 3 samples yields satisfactory results in few-shot settings while also reducing the size requirements of the target dataset.
We implement all methods on pre-trained UNet\cite{ronneberger2015u}, SwinUNet \cite{cao2022swin}, and MambaUNet \cite{wang2024mamba}, from Flemme \cite{zhang2024flemme} following the pre-training settings of
MONAI \cite{cardoso2022monai}.
\vspace{-0.2cm}
\subsection{Performance evaluation}
\vspace{-0.1cm}
\vspace{-0.1cm}
\begin{table*}[!]
\caption{Results of different domain adaptation strategies on MRI (left) and CT (right) datasets for three models. The segmentation evaluation metrics are the Dice score and the mean IoU score. Bold number: best score.}
\label{ASAP}
\vspace{-0.3cm}
    \begin{minipage}[t]{.5\linewidth}
        \centering
        \setlength\tabcolsep{4pt} 
        \footnotesize
     \resizebox{\linewidth}{!}{%
            \begin{tabular}{cccc|cc|cc}
                \toprule
                \multirow{2}{*}{Target} & \multirow{2}{*}{Method}
                 & \multicolumn{2}{c}{UNet} & \multicolumn{2}{c}{SwinUNet} & \multicolumn{2}{c}{MambaUNet} \\
                \cmidrule(lr){3-4}
                \cmidrule(lr){5-6}
                \cmidrule(lr){7-8}
                 & & \multicolumn{1}{c}{Dice (\%)} & \multicolumn{1}{c}{mIoU (\%)}
                 & \multicolumn{1}{c}{Dice (\%)} & \multicolumn{1}{c}{mIoU (\%)}
                 & \multicolumn{1}{c}{Dice (\%)} & \multicolumn{1}{c}{mIoU (\%)}\\
                \cmidrule(lr){1-8}
                \multirow{5}{*}{PVL} & Direct FT & 23.48 & 13.82 & 18.35 & 10.48 & 27.20 & 16.18\\ 
                 & GMS (NIPS' 20) & 23.50 & 13.56 & 20.41 & 11.86 & 27.45 & 16.72 \\
                 & MTL (MIA' 23) & 20.79 & 11.70 & 14.62 & 8.41 & 19.02 & 11.09 \\
                & DAL (NIPS' 24) & 26.76 & 15.99 & 19.65 & 11.26 & 26.92 & 16.23 \\ 
                \cmidrule{2-8}
                 & ASAP (Ours) & \textbf{27.95} & \textbf{16.80} & \textbf{22.16} & \textbf{13.02}
                 & \textbf{30.06} & \textbf{18.48}\\
                \cmidrule(lr){1-8}
                \cmidrule(lr){1-8}
                \multirow{5}{*}{WMH} & Direct FT & 57.03 & 40.78 & 64.24 & 48.79 & 62.35 & 45.51\\ 
                 & GMS (NIPS' 20) & 52.86 & 39.40 & 64.54 & 50.91 & 65.78 & 50.36 \\
                 & MTL (MIA' 23) & 58.13 & 41.29 & 53.69 & 38.83 & 53.81 & 38.76 \\
                & DAL (NIPS' 24) & 62.74 & 46.75 & 67.16 & 51.47 & 63.62 & 48.68 \\ 
                \cmidrule{2-8}
                 & ASAP (Ours) & \textbf{66.94} & \textbf{53.10} & \textbf{67.62} & \textbf{52.04}
                 & \textbf{71.06} & \textbf{56.04}\\
                 \bottomrule
                \end{tabular} }
    \end{minipage}
    \hfill
    \begin{minipage}[t]{.5\linewidth}
        \centering
        \setlength\tabcolsep{4pt} 
        \footnotesize
        \resizebox{\linewidth}{!}{
        \begin{tabular}{cccc|cc|cc}
        \toprule
        \multirow{2}{*}{Target} & \multirow{2}{*}{Method}
         & \multicolumn{2}{c}{UNet} & \multicolumn{2}{c}{SwinUNet} & \multicolumn{2}{c}{MambaUNet} \\
        \cmidrule(lr){3-4}
        \cmidrule(lr){5-6}
        \cmidrule(lr){7-8}
         & & \multicolumn{1}{c}{Dice (\%)} & \multicolumn{1}{c}{mIoU (\%)}
         & \multicolumn{1}{c}{Dice (\%)} & \multicolumn{1}{c}{mIoU (\%)}
         & \multicolumn{1}{c}{Dice (\%)} & \multicolumn{1}{c}{mIoU (\%)}\\
        \cmidrule(lr){1-8}
        \multirow{5}{*}{Vessel} & Direct FT & 48.70 & 33.13 & 41.31 & 26.64 & 49.08 & 33.39\\ 
         & GMS (NIPS' 20) & 48.74 & 33.85 & 36.16 & 22.51 & 46.62 & 40.94 \\
         & MTL (MIA' 23) & 22.64 & 13.12 & 32.23 & 19.58 & 35.05 & 21.86 \\
        & DAL (NIPS' 24) & 45.78 & 30.33 & 41.24 & 26.81 & 44.54 & 30.91 \\ 
        \cmidrule{2-8}
         & ASAP (Ours) & \textbf{49.22} & \textbf{33.55} & \textbf{48.05} & \textbf{32.81}
         & \textbf{49.27} & \textbf{33.34}\\
        \cmidrule(lr){1-8}
        \cmidrule(lr){1-8}
        \multirow{5}{*}{Liver} & Direct FT & 79.79 & 66.97 & 84.86 & 74.63 & 84.40 & 73.67\\ 
         & GMS (NIPS' 20) & 90.88 & 83.50 & 86.99 & 77.53 & 90.04 & 82.20 \\
         & MTL (MIA' 23) & 81.15 & 71.52 & 85.56 & 75.80 & 91.26 & 84.11 \\
        & DAL (NIPS' 24) & 91.66 & 84.22 & 84.06 & 73.05 & 87.71 & 78.50 \\ 
        \cmidrule{2-8}
         & ASAP (Ours) & \textbf{92.10} & \textbf{85.61} & \textbf{87.27} & \textbf{78.67}
         & \textbf{92.91} & \textbf{86.95}\\
         \bottomrule
        \end{tabular}
         }
    \end{minipage}
\vspace{-0.5cm}
\end{table*}
\begin{figure}[b]
\vspace{-0.5cm}
    \centering
    \includegraphics[width=\linewidth]{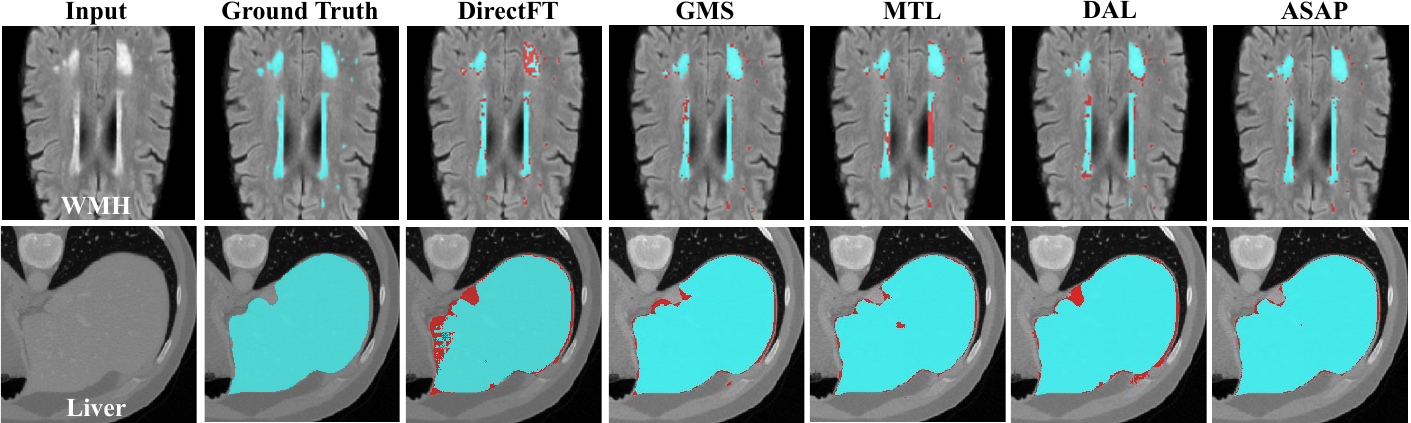}
    \vspace{-0.5cm}
    \caption{Visualization of different domain adaptation methods performance of two specific target tasks:
    WMH segmentation on MRI images and liver segmentation on CT images, both using MambaUNet. The pixels highlighted in red represent \textit{incorrect predictions}.
    }
    \label{result}
    \vspace{-0.5cm}
\end{figure}
We compare our framework with state-of-the-art few-shot domain adaptation methods: 
1) direct fine-tuning (FT) the source model on the target dataset,
2) GMS \cite{yu2020gradient} identifies one best auxiliary dataset to aid the target based on gradient magnitude similarity,
3) a mixed-batch multi-task learning (MTL) framework \cite{graham2023one} utilizes all auxiliary data simutaneously,
4) a dynamic auxiliary learning (DAL) method \cite{albalak2024improving} adaptively samples the auxiliary data to jointly train with the target dataset based on gradient alignment.
We evaluate the target segmentation performance using the Dice score and the mean IoU. 
A quantitative analysis of model adaptation performance on MRI and CT datasets is detailed in Table~\ref{ASAP}.
The proposed ASAP framework outperforms all the baselines on all datasets, across modalities and anatomical regions. 
We also present the WMH and liver segmentation results on MambaUNet of different methods, clearly demonstrating
the enhancements our method brings to the target few-shot medical image
segmentation tasks, as shown in Fig~\ref{result}.

\vspace{-0.4cm}
\subsubsection{Effectiveness of exploring and exploiting}
\vspace{-0.1cm}
The a-priori dataset selection method GMS is inferior to ours because it relies solely on exploiting the relations determined prior to training, but never exploring, e.g., as observed in the vessel experiment on the right side of Table~\ref{ASAP}.
In contrast, the multi-task learning (MTL) framework continuously explores all auxiliary data but never exploits knowledge of their relation to the target, leading to unsatisfactory results, e.g., as observed in the WMH experiment on the left side of Table~\ref{ASAP}, with significantly increasing training time--up to 34 times longer than direct fine-tuning.
By balancing the trade-off between exploration and exploitation, our ASAP achieves a 24.39\% gain on WMH compared to MTL, and a 13.18\% gain on vessel segmentation compared to GMS in Dice scores.
\vspace{-0.4cm}
\subsubsection{Effectiveness of the efficient reward function}
\vspace{-0.2cm}
Compared to static dataset selection methods GMS and MTL, DAL offers a relatively better solution by dynamically selecting the auxiliary data based on gradient alignment.
However, our reward function, with consideration of $\mathcal{R}^{PM}$ term, the reward of positive for model convergence, serves as a more effective guide to enable the model to 
converge faster and 
deliver superior performance.
Meanwhile, our reward function only relies on the losses and gradients, which are intrinsic to the model, making it naturally training-efficient: it took 15.28 hours to adapt the MambaUNet for the target task liver segmentation, compared to 75.03 hours in MTL, 15.97 hours in DAL, and 15.13 hours in GMS, the a-priori dataset selection method, with the same input size of 80 x 240 x 240 and batch size of 4.
As per our policy, we update only the selected arm's reward during training, which keeps the additional complexity stable, irrespective of the size of the auxiliary data pool. 

\vspace{-0.4cm}
\subsubsection{Investigating the active and sequential training dynamics.}
\begin{figure}[!]
\vspace{-0.1cm}
    \centering
    \includegraphics[width=\linewidth]{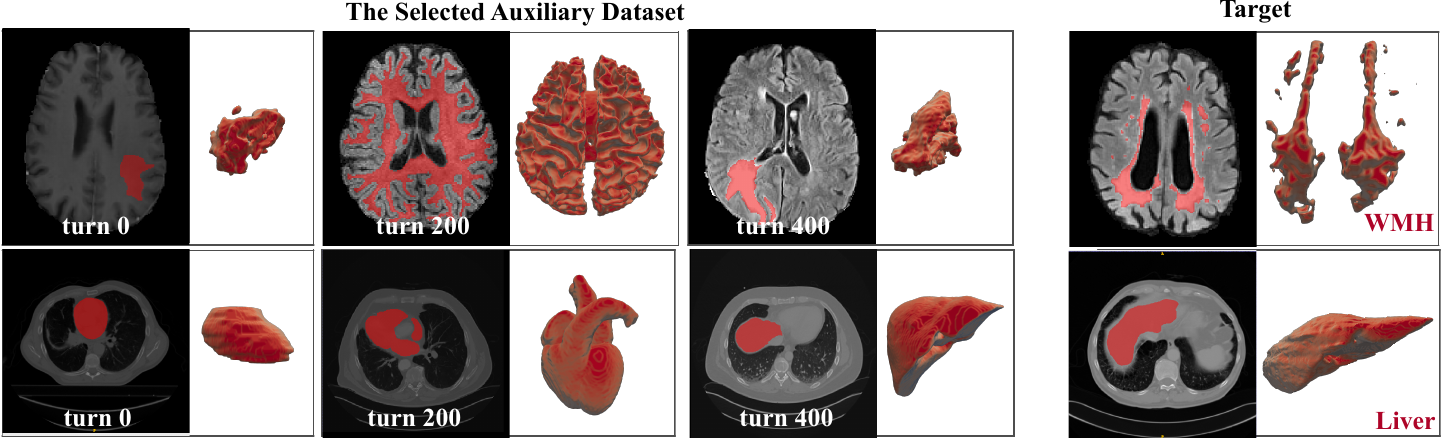}
    \vspace{-0.6cm}
    \caption{A 3D visualization of the active and sequential training process. The figure shows the selected auxiliary datasets at turn 0, turn 200, and turn 400, for two specific target tasks. The images are presented alongside their ground truth.
    }
    \label{selection}
    \vspace{-0.5cm}
\end{figure}

\vspace{-0.2cm}
A closer look at the selected auxiliary tasks illustrates the active and sequential adaptation training mechanism, visualized in Fig~\ref{selection}.
We show the selected auxiliary datasets at different turns for target tasks WMH segmentation on MRI images and liver segmentation on CT images.
Interestingly, the policy does not initially sample the task experientially similar to the target. Instead, it sequentially selects the auxiliary dataset that progressively aligns with the target.
Despite lacking access to the source domain, we can still effectively narrow the domain discrepancy by following this step-by-step knowledge acquisition, demonstrating the strength of the active and sequential domain adaptation framework.
\vspace{-0.4cm}
\section{Conclusion}
\vspace{-0.3cm}
We propose a novel active and sequential domain adaptation (ASAP) framework to adapt foundation models for the few-shot medical image segmentation tasks. 
With our desiderata in mind, the proposed ASAP achieve:
1) no requirement for access to the source domain or a substantial amount of target data,
2) incorporation of auxiliary data with dynamic scheduling of prioritized learning, adding minimal extra memory and computational overhead,
3) effective and efficient adaptation of foundational models, leading to strong performance on the target task.
We believe our proposed approach will better leverage public medical resources, including foundation models and available auxiliary datasets, to tailor a model for the desired few-shot target task in a fast and scalable way. 



\vspace{-0.3cm}
\section{Compliance with ethical standards}
\vspace{-0.3cm}
\label{sec:ethics}


This research study was conducted retrospectively using the animal subject data made available in open access by 
\cite{pati2021federated,sun2021multi,kuijf2019standardized,wasserthal2023totalsegmentator,antonelli2022medical}.
Ethical approval was not required as confirmed by the license attached with the open access data.
\vspace{-0.4cm}

\bibliographystyle{IEEEbib}
\bibliography{main}

\end{document}